\documentclass{article}

\usepackage[preprint]{nips_2018}

\usepackage[utf8]{inputenc} % allow utf-8 input
\usepackage[T1]{fontenc}    % use 8-bit T1 fonts
\usepackage{hyperref}       % hyperlinks
\usepackage{url}            % simple URL typesetting
\usepackage{booktabs}       % professional-quality tables
\usepackage{amsfonts}       % blackboard math symbols
\usepackage{nicefrac}       % compact symbols for 1/2, etc.
\usepackage{microtype}      % microtypography
\usepackage{natbib}         % natbib for references
\usepackage{graphicx}
\usepackage{appendix}
\usepackage[ruled]{algorithm2e}
\usepackage{subcaption}
\graphicspath{ {images/} }

\usepackage{multirow}
\usepackage{mathtools}
\usepackage{verbatimbox}
\usepackage{tkz-graph}
\usepackage{siunitx}
\usepackage[printfigures]{figcaps}
\usetikzlibrary{positioning}
\tikzset{main node/.style={circle,fill=blue!20,draw,minimum size=1cm,inner sep=0pt},}

\newcommand{\neno}{\newline\noindent} 

\DeclareMathOperator*{\argminA}{arg\,min}

\title{Regression by clustering using Metropolis-Hastings}

\author{
      Simón~Ramírez-Amaya \\
      Universidad de los Andes \\
      Bogota, Colombia \\
      \texttt{s.ramirez34@uniandes.edu.co} \\
      \And
      Adolfo J.~Quiroz \\
      Universidad de los Andes \\
      Bogota, Colombia \\
      \texttt{aj.quiroz1079@uniandes.edu.co} \\
      \And
      Álvaro~Riascos \\
      Quantil and Universidad de los Andes \\
      Bogota, Colombia \\
      \texttt{alvaro.riascos@quantil.com.co} \\
}

\begin{document}

\maketitle
  
\begin{abstract}
    High quality risk adjustment in health insurance markets weakens insurer incentives to engage in inefficient behavior to attract lower-cost enrollees. We propose a novel methodology based on Markov Chain Monte Carlo methods to improve risk adjustment by clustering diagnostic codes into risk groups optimal for health expenditure prediction. We test the performance of our methodology against common alternatives using panel data from 500 thousand enrollees of the Colombian Healthcare System. Results show that our methodology outperforms common alternatives and suggest that it has potential to improve access to quality healthcare for the chronically ill.   
\end{abstract}

\section{Introduction}

Contrary to expenditures on other services, health care expenditures are characterized both by large random variation as well as large predictable variation across individuals \citep{VandeVenchapter}. Such differences create potential for efficiency gains due to risk reduction from insurance and raise concerns about fairness across individuals with different expected needs.

However, widespread health insurance under a uniform pricing restriction creates an important tradeoff between \textit{efficiency in production} and \textit{selection} \citep{Newhouse}. By \textit{efficiency in production} we mean least cost medical treatment of a medical problem, holding quality constant. By \textit{selection} we mean actions of agents on either side of the market to benefit from unpriced risk heterogeneity. A uniform fully prospective payment to producers yields efficient production because the firm captures any surplus. However, uniform fully prospective payment for a heterogenous group of persons gives the firm maximum incentives to select good risks and avoid bad ones (see Figure 1). 

Subsidies based on the observable characteristics of consumers have become an increasingly important regulatory tool. These type of subsidies smooth the uniform pricing restriction that gives rise to selection incentives under prospective payment. From the producer perspective, these subsidies have the potential to decouple expected costs from expected profit and thus weaken incentives to manipulate insurance products to attract lower-cost consumers. Consumer-based subsidies to insurers are known as risk adjustment, and their introduction has been motivated by a broader shift towards regulated private insurance markets \citep{GerusoLayton,Gruber}.

\begin{figure}
    \centering
    \caption{Relaxation of the uniform pricing restriction and the \textit{selection} and \textit{efficiency in production} trade-off}
    \begin{subfigure}[b]{0.55\textwidth}
       \begin{tikzpicture}
            \draw [thin, gray, ->] (0,-0.5) -- (0,4)      % draw y-axis line
                node [left, black] {$inefficiency$};              % add label for y-axis
        
            \draw [thin, gray, ->] (-0.5,0) -- (4,0)      % draw x-axis line
                node [below, black] {$selection$};              % add label for x-axis
        
            \draw [draw=black,thick] (0,3.75) -- (3.75,0);% draw the graph
            
            \draw [dotted,draw=red,ultra thick] (0,0) -- (4,4);% draw the graph
            
            \draw [dotted,draw=black,thick] (0,1) -- (2.75,1);% draw the graph
            
            \draw [dotted,draw=black,thick] (2.75,0) -- (2.75,1);% draw the graph
            
            \draw [dotted,draw=black,thick] (1,0) -- (1,2.75);% draw the graph
            
            \draw [dotted,draw=black,thick] (0,2.75) -- (1,2.75);% draw the graph
        
            \node [right] at (0.75,3.75) {\textcolor{red}{fee-for-service}};                % label y-intercept
            \node [right] at (3.0,1.5) {\textcolor{red}{prospective payment}};               % label x-intercept
        \end{tikzpicture}
       \caption{Trade-off under uniform pricing}
       \label{fig:Ng1} 
    \end{subfigure}
    
    \begin{subfigure}[b]{0.55\textwidth}
       \begin{tikzpicture}
            \draw [thin, gray, ->] (0,-0.5) -- (0,4)      % draw y-axis line
                node [left, black] {$inefficiency$};              % add label for y-axis
        
            \draw [thin, gray, ->] (-0.5,0) -- (4,0)      % draw x-axis line
                node [below, black] {$selection$};              % add label for x-axis
        
            \draw [dotted,draw=black,thick] (0,3.75) -- (3.75,0);% draw the graph
            
            \draw [draw=black,thick] (0,3.75) -- (2.75,0);% draw the graph
            
            \draw [dotted,draw=red,ultra thick] (0,0) -- (4,4);% draw the graph
            
            \draw [dotted,draw=black,thick] (0,1) -- (2.75,1);% draw the graph
            
            \draw [dotted,draw=black,thick] (2.75,0) -- (2.75,1);% draw the graph
            
            \draw [dotted,draw=black,thick] (1,0) -- (1,2.75);% draw the graph
            
            \draw [dotted,draw=black,thick] (0,2.75) -- (1,2.75);% draw the graph
        
            \node [right] at (0.75,3.75) {\textcolor{red}{fee-for-service}};                % label y-intercept
            \node [right] at (3.0,1.5) {\textcolor{red}{prospective payment}};               % label x-intercept
        \end{tikzpicture}
       \caption{Trade-off under risk adjustment}
       \label{fig:Ng2}
    \end{subfigure}
    \begin{subfigure}[b]{0.55\textwidth}
       \begin{tikzpicture}
            \draw [thin, gray, ->] (0,-0.5) -- (0,4)      % draw y-axis line
                node [left, black] {$inefficiency$};              % add label for y-axis
        
            \draw [thin, gray, ->] (-0.5,0) -- (4,0)      % draw x-axis line
                node [below, black] {$selection$};              % add label for x-axis
        
            \draw [dotted,draw=black,thick] (0,3.75) -- (3.75,0);% draw the graph
            
            \draw [dotted,draw=black,thick] (0,3.75) -- (2.75,0);% draw the graph
            
            \draw [draw=black,thick] (0,3.75) -- (2.0,0);% draw the graph
            
            \draw [dotted,draw=red,ultra thick] (0,0) -- (4,4);% draw the graph
            
            \draw [dotted,draw=black,thick] (0,1) -- (2.75,1);% draw the graph
            
            \draw [dotted,draw=black,thick] (2.75,0) -- (2.75,1);% draw the graph
            
            \draw [dotted,draw=black,thick] (1,0) -- (1,2.75);% draw the graph
            
            \draw [dotted,draw=black,thick] (0,2.75) -- (1,2.75);% draw the graph
        
            \node [right] at (0.75,3.75) {\textcolor{red}{fee-for-service}};                % label y-intercept
            \node [right] at (3.0,1.5) {\textcolor{red}{prospective payment}};               % label x-intercept
        \end{tikzpicture}
       \caption{Trade-off under high quality risk adjustment}
       \label{fig:Ng2}
    \end{subfigure}
\end{figure}

Several policy choices need to be fine-tuned for risk-adjustment systems to work properly. Among them is estimation of risk-adjusted payments. Since the cost level of a health care service package is hard to determine, payments are based on observed expenses rather than needs-based costs \citep{VandeVenchapter}. High quality estimation is vital since poor risk adjustment gives insurance companies incentives to engage in cream-skimming \citep{Cutler,VandeVenarticle,Shmueli}. This type of behavior threatens risk solidarity, efficiency and possibly the unraveling of the insurance market itself.

Evidence suggests that demographic adjusters such as age, sex and residence are weak predictors of individual expenditure and that risk adjustment can be greatly improved by using diagnosis-based information  \citep{VandeVenchapter}. Since the early 1980's a considerable amount of research has developed risk adjustment models that use diagnostic from insurance claims to estimate risk-adjusted payments. 

Although each model has its own unique features, they share two characteristics that are worth highlighting. First, all models rely on the diagnostic standard known as the International Classification of Diseases (ICD) by the World Health Organization (WHO). In its 10th revision, the ICD allows for more than $16,000$ different codes for disease identification. Second, since the code space size is not trivial, models rely on classification systems to cluster ICD codes into meaningful Diagnostic Related Groups (DRG).  

Traditionally, DRG have been constructed using \textit{ad hoc} expert criteria on clinical, cost and incentive considerations \citep{VandeVenchapter,Juhnke}. The most refined versions of these classification systems begin by classifying diagnoses into a tractable number of diagnostic-based groups and then using these diagnostic groups to classify individuals according to the specific combination of conditions each individual has. More recently, there have been attempts to inform DRG construction by introducing iterative hypothesis testing \citep{3M}. 

Despite the wealth of classification systems available, quality of risk adjustment remains limited \citep{Kleef,Brown,Alfonso,Riascos}. We believe that risk adjusment can be improved by formally approaching the problem of finding optimal DRG. In this paper we develop a methodology aimed at solving this problem by using Monte Carlo Markov Chain (MCMC) methods to efficiently traverse the space of possible solutions. We also test our methodology against common alternatives in the Colombian Health Sector using two year panel data for 3.5 million enrollees.  Results show that our methodology outperforms common alternatives and suggest that it has potential to improve access to quality healthcare for the chronically ill.

The remainder of this paper is organized as follows. Section 2 provides some background. Section 3 introduces the theoretical framework designed to approach the problem of finding optimal DRG. Section 4 describes the empirical framework and main results. Finally, section 5 concludes and outlines directions for future work.

\section{Background}

\subsection{ICD-10}

A classification of diseases is a system of categories to which morbid entities are assigned according to established criteria \citep{ICD10}. The purpose of the ICD is to permit systematic recording of mortality and morbidity data collected across countries and time. In practice, the ICD is used to translate diagnoses and health problems into alphanumeric codes which allows easy storage, retrieval and analysis of the data. The ICD is the most widely used diagnostic classification system and has become the international standard for all general epidemiological and many health-management purposes in clinical, administrative and research activities.

The basic ICD is a single coded list of three-character categories, each of which can be further divided into up to 10 four-character subcategories. The 10th revision (ICD-10) uses an alphanumeric code with a letter in the first position and a number in the second, third and fourth positions. The fourth character follows a decimal point. Possible code numbers therefore range from \texttt{A00.0} to \texttt{Z99.9}. Codes \texttt{U00} to \texttt{U99} are unused since they are reserved for the provisional assignment of new diseases of uncertain etiology and for the testing of alternative classifications in research projects. 

The core classification, that is, the list of three-character categories, is the mandatory level for reporting to the WHO mortality database and for general international comparisons. Some of the three-character categories are for single conditions, selected because of their frequency, severity or susceptibility to public health intervention, while others are for groups of diseases with some common characteristic. 

The ICD-10 allow for different levels of detail by grouping three-character categories into a hierarchical structure. Three-character categories are grouped into semantically relevant \texttt{blocks} of variable length. For example, three-character categories starting with letter A or letter B are grouped into 21 blocks reflecting two axes of classification: mode of transmission and broad group of infecting organisms. Block \texttt{B65-B83} groups three-character categories related to \texttt{Helminthiases} while block \texttt{B85-B89} covers \texttt{Pediculosis, acariasis and other infestations}.

Blocks are further grouped into 22 overarching \textit{chapters}. Each chapter contains sufficient three-character categories to cover its content. Not all available codes are used ($2,048$ out of $2,600$ possible three-character categories), allowing space for future revision and expansion. The first character of the ICD code is a letter, and each letter is associated with a particular chapter, except for the letter D, which is used in both Chapter II, \texttt{Neoplasms}, and Chapter III, \texttt{Diseases of the blood and blood-forming organs and certain disorders involving the immune mechanism}, and the letter H, which is used in both Chapter VII, \texttt{Diseases of the eye and adnexa} and Chapter VIII, \texttt{Diseases of the ear and mastoid process}. Four chapters (Chapters I, II, XIX and XX) use more than one letter in the first position of their codes.  

It is important to note that ICD-10 offers no one-to-one correspondence between one-character or two-character codes to meaningful medical classifications. Any attempt to group ICD-10 diagnostic codes into DRG should take place in the space of meaningful medical classifications: chapters, blocks and three-or-more-character categories. 

According to \cite{ICD10} the ICD has developed as a practical, rather than a purely theoretical classification, in which there are a number of compromises between classification based on etiology, anatomical site, circumstances of onset, state of knowledge, etc. There have also been adjustments to meet the variety of statistical applications for which the ICD is designed, such as mortality, morbidity, social security and other types of health statistics and surveys.

\subsection{Colombian Health Sector}

In Colombia, law 100 of 1993, transformed the public health system into a competitive insurance market \citep{Riascos}. This market structure has five key components. First, the Colombian health sector comprises a contributory and a subsidized system. All formal employees and their beneficiaries are enrolled to the former, while people with no sources of income and who are poor enough to qualify are enrolled to the latter. Of the 46 million people enrolled, 44\% are in the contributory system and 56\% in the subsidized system.  Second, a wide-ranging benefits package that defines all the services enrollees have the right to claim. Third, a group of private sector health insurers (EPS) who enroll population on a one-to-many basis and configure a network of health service providers (IPS) in charge of delivering services. Fourth, universal open enrolment. Fith and last, a mechanism for the payment of such services that controls for risk heterogeneity across enrollees. Monthly mandatory risk premium fees are collected from all employed enrollees and then redistributed as capitation payments to health insurers in the market along with additional tax related funding. 

Capitation payments are payed on a subscription basis and are adjusted by each enrollee health risk. In the absence of risk-adjusted payments, health insurers have perverse incentives to engage in risk selection. For example, they would discourage the enrollment of high-risk individuals through strenuous formalities, large waiting lines, or unobservable low service quality \citep{CASTANO}. This payment scheme serves the double purpose of configuring a cross-subsidies system that helps insurers mitigate their financial risk and reduce the incentives to “\textit{cream skim}” and being a expenditure containment mechanism  \citep{VandeVenarticle}.

Although there is great uncertainty revolving annual health expenditures, a part of it is predictable by socio-demographic variables. For example, women in childbearing age are costlier than men in the same age group and elders are costlier than teenagers. In Colombia, risk-adjustment is based on the risk pools formed by unique combinations of gender, age group and enrollee’s location. Currently the mechanism consists of a linear regression of annual health expenditure on sociodemographic risk factors. Age groups are defined by the Ministry of Health and location is a categorization of the municipality of residence in three areas: urban (metropolitan areas), normal (municipalities surrounding metropolitan areas) and special (peripheral municipalities). However, the socio-demographic characteristics of individuals only explain 2\% of the variation in health expenditure \citep{Riascos}.

\section{Statistical Framework}

We introduce a novel statistical framework based on the Metropolis-Hastings algorithm \citep{hastings70} aimed at finding optimal groupings of categorical variables in prediction problems. Whenever possible, we will provide additional insight into its immediate application to the risk-adjustment problem. That is, finding risk groupings of the ICD-10 diagnostic category space that minimize health expenditure prediction error. However, it should be kept in mind that our statistical framework can be applied to solving prediction problems not related to risk management in healthcare insurance markets.

We will begin by introducing the mathematical notion of the \textit{partition of a set} and its importance to our setup. We will build upon this notion to define a precise optimization problem. Finally we propose a solution based on the Metropolis-Hastings algorithm with particular transition dynamics.

\subsection{Partitions}

A \textit{partition} of size $k$ of a set $S$ is a grouping of the elements of $S$ into $k$ identifiable subsets in such a way that every element is included in one and only one of the subsets. Figure 2 (a) shows an example of a partition o size $2$ over the set $S=\{s_{1},s_{2},s_{3}\}$. In this trivial example, element $s_{1}$ is assigned to group $k_{1}$, while elements $s_{2}$ and $s_{3}$ are assigned to group $k_{2}$.

This notion is very useful. If we model the space of ICD-10 diagnostic categories as a set, then a partition of this set of size $k$ effectively models a possible risk grouping into $k$ different risk groups.

\subsection{Problem}
Let $I$ be a finite index over a set of observations $\tau = \{(x_{i},c_{i},y_{i})\}_{i \in I}$ drawn from an arbitrary joint distribution $\Omega$ over $X \times C \times Y$. $x_{i}$ is a vector of characteristics with index $i \in I$ that compromises continuous, and discrete variables, $c_{i}$ is a categorical variable and $y_{i}$ is a continuous real dependent variable. Let $X$ be the set of continuous and discrete characteristics, $C$ be the set of categorical characteristics and $\mathbb{R}$ the set of real numbers. $n=\mid C \mid$ is large. \\

We want to learn a hypothesis $f: X \times C \rightarrow \mathbb{R}$ that minimizes a loss function $L:\mathbb{R} \times \mathbb{R}\rightarrow \mathbb{R}$, where $L(y,\hat{y})$ is the loss associated to an example (x,c,y) and the prediction $\hat{y}=f(x,c)$. We will do this using as plausible hypothesis a set of local linear functions. That is, a linear function of $x$ is used on those examples in which $c$ falls in a given subset of $C$. We expect to improve performance by reducing the dimension of the categorical feature space. Let $P$ the set of all partitions of $C$, fix a natural number $k$ and let $P_{k}$ be the set of all partitions of $C$ into $k$ clusters. Let $\mathcal{H}$ be the set of all locally linear hypothesis that use a linear predictor on each cluster defined by $p$ $\in P_k$. For every partition $p$ we can define an optimal learning hypothesis $h_{p} \in \mathcal{H}$.

For a fixed $k$ the problem we want to solve is:

\begin{equation*}
    \min_{p \in P_{k}} \mathcal{L}(p) = E[L(y,h_{p}(x,c))]
\end{equation*}

The expected loss for a particular choice of the hypothesis $h_p$, associated to a partition, can be estimated by its empirical counterpart on  a large sample.

This optimization setup fits gracefully the risk-adjustment problem. In this case set $C$ models the set of ICD-10 categories. $P_{k}$ is the space of all partitions of diagnostic categories of size $k$. $Y$ models health expenditure and $X$ a set of demographic covariates. Index $i$ indexes consumers for which triplets $(x_{i},c_{i},y_{i})$ are observed. For a fixed $k$, the risk adjustment problem consists of finding a partition $p$ from $P_{k}$ that minimizes a loss function that compares observed expenditure against a linear hypothesis constructed using $p$ and demographic covariates $x$.

\subsection{Solution}

Our objective function is discrete, non-trivial and has a huge candidate space.  Our problem seems a good candidate for the stochastic optimization approach \citep{rennard}. Loosely put, the stochastic approach consists of picking candidates $p_{1},...,p_{n}$ randomly from $P_{k}$ using some sampling distribution and considering $\argminA_{i\in n} \mathcal{L}(p_{i})$ as the solution. We are interested in making probability under the sampling distribution proportional to performance under the objective function.

The \textit{normalized exponential function} allows us to create a probability distribution over partitions from their non-normalized performance under $\mathcal{L}$:
        
        \begin{equation*}
            \pi(p)=\frac{exp(\frac{-\mathcal{L}(p)}{T})}{Z}
        \end{equation*}
        
where $T$ is a parameter known as \textit{temperature} and $Z$ is the \textit{normalization constant}:
        
        \begin{equation*}
            Z = \sum_{q \in P_{k}}exp(\frac{-\mathcal{L}(q)}{T})
        \end{equation*}
        
Sampling distribution $\pi(p)$ is known up to $Z$ and therefore sampling directly from $\pi$ is not possible. However, we can construct a stochastic model to sample from $\pi(p)$ in the long run. Loosely put, this stochastic model is simply a graph-like structure over $P_{k}$ where we perform a particular type of random walk specified by the Metropolis-Hastings algorithm. The idea is to use an approximation to the expected loss, associated to each partition as the "energy" function in a Metropolis-Hastings, or  Simulated Annealing algorithm, see \cite{bertsimas1993}, for details. Simulated Annealing is the "changing temperature" version of the Metropolis-Hastings algorithm. In these procedures, calculation of the normalization constant $Z$ is never required, since it cancels out in the transition probability. In the remainder of this section we explain the details of transition dynamics in our solution to the optimal partitioning problem. 

\subsubsection{Distance between partitions}

In order to use the Metropolis-Hastings algorithm we first need to introduce a notion of similarity among partitions that informs the local exploration process. We will use the distance proposed by \cite{Gusfield} described next (see also \cite{Rossi}). Given $p,q \in P_{k}$, define the partition distance, $D(p,q)$ between $p$ and $q$, as the minimum number of elements that must be deleted from $C$ so that the restrictions of $p,q$ to the remaining elements are identical. Equivalently, $D(p,q)$ is the minimum number of elements that need to be reassigned to a different cluster in $p$. This last interpretation is the most intuitive in the context of the Metropolis-Hastings algorithm. Figure 2 provides a trivial example of distance between partitions.

\begin{figure*}
    \setcounter{subfigure}{0}
        \centering
        \begin{subfigure}[b]{0.475\textwidth}
            \centering
                \begin{tikzpicture}
                    \node[main node] (1) {$s_{1}$};
                    \node[main node] (2) [below = 1.0cm of 1]  {$s_{2}$};
                    \node[main node] (3) [below = 1.0cm of 2]  {$s_{3}$};
                    
                    \node[main node] (4) [below right = 0.5cm and 1.5cm of 1] {$k_{1}$};
                    \node[main node] (5) [above right = 0.5cm and 1.5cm of 3]  {$k_{2}$};
                    
                    \path[draw,thick]
                    (1) edge node {} (4)
                    (2) edge node {} (5)
                    (3) edge node {} (5);
                \end{tikzpicture}
            \caption[Network2]%
            {{\small Partition $p$ over $S$}}    
            \label{fig:mean and std of net14}
        \end{subfigure}
        \hfill
        \begin{subfigure}[b]{0.475\textwidth}  
            \centering 
            \begin{tikzpicture}
                    \node[main node] (1) {$s_{1}$};
                    \node[main node] (2) [below = 1.0cm of 1]  {$s_{2}$};
                    \node[main node] (3) [below = 1.0cm of 2]  {$s_{3}$};
                    
                    \node[main node] (4) [below right = 0.5cm and 1.5cm of 1] {$k_{1}$};
                    \node[main node] (5) [above right = 0.5cm and 1.5cm of 3]  {$k_{2}$};
                    
                    \path[draw,thick]
                    (1) edge node {} (4)
                    (2) edge node {} (4)
                    (3) edge node {} (5);
                \end{tikzpicture}
            \caption[]%
            {{\small Partition $q$ over $S$}}    
            \label{fig:mean and std of net24}
        \end{subfigure}
        \vskip\baselineskip
        \begin{subfigure}[b]{0.475\textwidth}   
            \centering 
            \begin{tikzpicture}
                    \node[main node] (1) {$s_{1}$};
                    \node[main node] (3) [below = 1.0cm of 2]  {$s_{3}$};
                    
                    \node[main node] (4) [below right = 0.5cm and 1.5cm of 1] {$k_{1}$};
                    \node[main node] (5) [above right = 0.5cm and 1.5cm of 3]  {$k_{2}$};
                    
                    \path[draw,thick]
                    (1) edge node {} (4)
                    (3) edge node {} (5);
                \end{tikzpicture}
            \caption[]%
            {{\small Partition $p^\prime$ over $S \setminus \{s_{2}\}$}}   
            \label{fig:mean and std of net34}
        \end{subfigure}
        \quad
        \begin{subfigure}[b]{0.475\textwidth}   
            \centering 
            \begin{tikzpicture}
                    \node[main node] (1) {$s_{1}$};
                    \node[main node] (3) [below = 1.0cm of 2]  {$s_{3}$};
                    
                    \node[main node] (4) [below right = 0.5cm and 1.5cm of 1] {$k_{1}$};
                    \node[main node] (5) [above right = 0.5cm and 1.5cm of 3]  {$k_{2}$};
                    
                    \path[draw,thick]
                    (1) edge node {} (4)
                    (3) edge node {} (5);
                \end{tikzpicture}
            \caption[]%
            {{Partition $q^\prime$ over $S \setminus \{s_{2}\}$}}     
            \label{fig:mean and std of net44}
        \end{subfigure}
        \caption[ The average and standard deviation of critical parameters ]
        \centering
        {\small Example over a set of elements $C=\{c_{1},c_{2},c_{3}\}$ and groups $K=\{k_{1},k_{2}\}$. Distance between partitions $p$ and $q$ is  $1$ since $p$ is not equal to $q$ and $p^\prime$ is equal to $q^\prime$.}
        \label{fig:mean and std of nets}
    \end{figure*}

\subsubsection{Counting partitions}

We are interested in counting neighboring partitions in order to specify well suited transition distributions among partitions. Let $p \in P_{k}$ be an arbitrary partition of size $k$ of the space of categorical characteristics $C$ where $n = |C|$. How many partitions of size $k$ at distance at most $j$ exist from partition $p$? 

It is trivial to see that $p$ is the only partition at distance $0$ from itself: 
$$ |\{q \in P_{k} | D(p,q) = 0\}| = 1 $$ 

Each of the $j$ chosen elements in $C$ can be assigned to $k-1$  different clusters so that the resulting partition $q$ is exactly at distance $1$ from $p$: 
$$ |\{q \in P_{k} | D(p,q) = 1\}| = n(k-1) $$ 

Furthermore, each element in every pair of distinct elements in $C$ can be reassigned in $(k-1)$ different clusters so that the resulting partition $q$ is exactly at distance $2$ from $p$: 
$$ |\{q \in P_{k} | D(p,q) = 2\}| = \frac{n(n-1)}{2}(k-1)^{2} $$ 

The preceding reasoning can be generalized. The number of ways in which $j$ distinct elements can be chosen out of $n$ is: 
$$ {n\choose j} = \frac{n!}{j!(n-j)!} $$ 

Each element $0,...,j-1$ can be assigned to $k-1$ different clusters. Therefore:

$$N_{p,j}=|\{q \in P_{k} | d(p,q) = j\}| = 
{n\choose j}(k-1)^{j}$$ 

$$N_{p,\leq j}=|\{q \in P_{k} | d(p,q) \leq j\}| = \sum\limits_{i=0}^{j}{n\choose i}(k-1)^{i}$$

In order to produce new partitions in $P_k$, the procedure just described, of reassignment of elements in $C$, requires the groups in the original partition $p$ to be large enough. For instance, if $p$ includes clusters of size 1 or 2, then reassignment of 1 or 2 elements in those clusters could result in the elimination of a cluster and reduce the number of non empty parts in the resulting partition. We have allowed for parts in our implementation to be temporarily empty, since it happens very rarely, but a rejection procedure could be used to produce a random walk in $P_k$, as follows: 
\neno (i) Select a new partition $q$ by reassignment of $j$ elements following the 
formula given above.
\neno(ii) If the new partition chosen has less than $k$ parts, reject it and
go back to step (i). Otherwise accept it as the candidate new partition.
\neno Since under uniform sampling, the conditional distributions are uniform,the rejection procedure will produce a partition $q$ uniformly chosen among those in $P_k$ at distance $j$ from $p$ that can be obtained by reassignment of elements to the existing clusters. For this procedure to be practical, that is, to have
a low rejection probability, the value of $j$ should be kept small.  

\subsubsection{Transition distributions and partition counts}
We can use the results from the preceding section to specify transition distributions $Q(p,q)$ over $P_{k}$ at will. For example, the uniform transition probability over all partitions at distance at most $j$:         \[
            Q(p,q) =
            \begin{cases}
                \frac{1}{\sum\limits_{i=0}^{j} {n\choose i}(k-1)^{i}} & \text{if } D(p,q) \leq j \\
                0        & \text{otherwise}
            \end{cases}
        \] In the special case $j=n$, this is the uniform transition distribution over all partitions.
        
We can use the counting results to specify more complex transition distributions as well. Suppose $j$ is not deterministic but random with some known discrete distribution $\mathcal{J}(\cdot)$ over the set \{$0,...,n\}$. The following distribution allocates even shares of probability mass $P[\mathcal{J}(\cdot)=D(p,q)]$ among $\{q \in P_{k} | D(p,q) = j\}$ for all $D(p,q) \in \{0,...,n\}$:

\[
            Q(p,q) =
                \frac{1}{{n\choose D(p,q)}(k-1)^{D(p,q)}}P[\mathcal{J}(\cdot)=D(p,q)]\\
        \]

Sampling from these probability distributions is not trivial since partitions counts can get very big quickly. In Appendix A we propose an algorithm to sample from these distributions without explicitly calculating counts and prove its correctness.

\subsubsection{Acceptance ratio}

Any transition distribution $Q(p,q)$ can be used to produce a Markov chain that has $\pi$ as the stationary distribution. Let the acceptance ratio $\alpha: P_{k} \times P_{k} \rightarrow \mathbb{R}$ be: $$\alpha(p,q)=\frac{\pi(q)Q(q,p)}{\pi(p)Q(p,q)}$$ and define:  
\begin{equation*}
    K(p,q) =
    \begin{cases*}
      Q(p,q) & if $p \neq q, \alpha(p,q) \geq 1$ \\
			Q(p,q)\alpha(p,q) & if $p \neq q, \alpha(p,q) < 1$\\
      Q(p,q) \quad + \\ \quad \sum\limits_{r:\alpha(p,r)<1}Q(x,r)(1-\alpha(x,r))        & otherwise
    \end{cases*}
\end{equation*}

Note that the normalization constant $Z$ is irrelevant for the definition of $K(p,q)$ and therefore the acceptance ratio can be calculated at each iteration of the algorithm. It is a well stated result that $\pi K = \pi$ and for any $p_{0}$ initial partition, $K^{n}(p_{0},p) \rightarrow \pi(p)$\citep{hastings70}. 

\section{Empirical Framework}

\subsection{Data}

We work with a two-year panel (2010 and 2011) containing a sample of $500,000$ enrollees to the contributive system who claimed at least one service in the first year. For every enrollee we observe year-wise-socio demographic variables. Gender is a binary variable, age groups are defined by the Ministry of Health and location is a categorization of the municipality of residence in three areas: urban (metropolitan areas), normal (municipalities surrounding metropolitan areas) and special (peripheral municipalities).

We also observe year-wise binary diagnosis variables for every three-character ICD-10 category. Each of these dummy variables indicate whether an enrollee claimed at least one service associated to the relevant diagnostic code as reported by the insurer. Finally, we also observe year-wise aggregate expenditure by enrollee as reported by insurers. 

The distribution of socio-demographic variables in our our sample resembles country-wide available information. Figure 3 compares group age distribution in our sample in 2011 against projected values based on the 2005 National Census \citep{DANE}. Comparison suggests our sample has a slight over representation of working age groups. This is probably related to the fact that our sample is drawn from population in the contributory system who actually claimed services.  Regarding gender, our sample has $55.6\%$ prevalence of females while the projection for 2011 was $50.6\%$. There are no country-wide estimates of residence for 2011. Our sample prevalence of metropolitan areas is $53.6\%$ while municipal and peripheral prevalence are $43.5\%$ and $2.9\%$ respectively. 

\begin{figure}
    \centering
    \caption{Density of age groups}
    \includegraphics[scale=0.5]{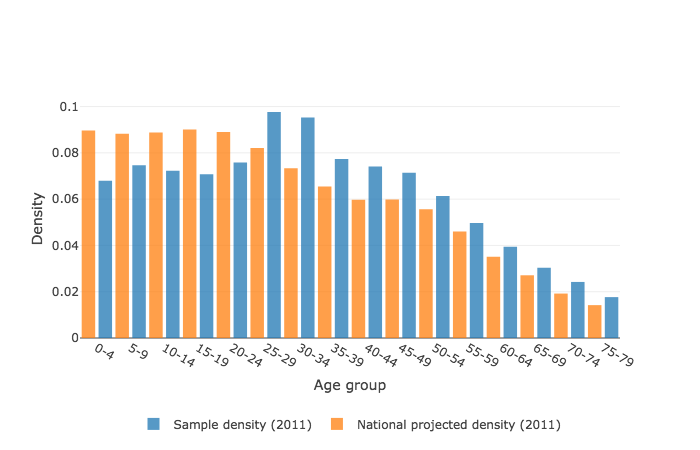}
\end{figure}

Out of the $2,048$ three-character categories in ICD-10, $1,868$ have at least one positive diagnosis in our sample. Figure 4 shows the distribution of these set of categories. $995$ categories have at least $50$ positive diagnoses. We will denote this set of three-character categories as \textit{observed}. We choose this thresholded set of categories as our working set in order to guarantee that for any reasonable random split we will have positive diagnoses in both train and test sets for each category. They are mostly related to low-prevalence diseases (e.g. \texttt{N27 Small kidney of unknown cause}), diseases in the verge of erradication (e.g. \texttt{B72 Dracunculiasis}) and highly specific conditions included in ICD-10 for the sake of completeness (e.g. \texttt{X52 Prolonged stay in weightless environment}).

\begin{figure}
    \centering
    \caption{Distribution of diagnosis frequency over three-character categories in 2010}
    \includegraphics[scale=0.5]{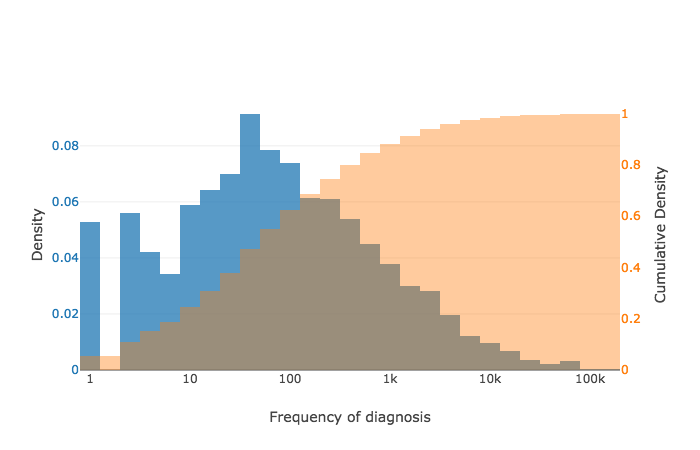}
\end{figure}

Mean expenditure varies significantly across positive diagnosis groups, suggesting three-character categories are powerful separators of expenditure. Figure 5 shows the distribution of mean expenditure over the observed three-character categories and Figure 6 shows word clouds constructed from the categories description for the upper and lower tails 5\% of the expenditure distribution. 

\begin{figure}
    \centering
    \caption{Distribution of mean expenditure in 2011 over observed three character categories in 2010 }
    \includegraphics[scale=0.5]{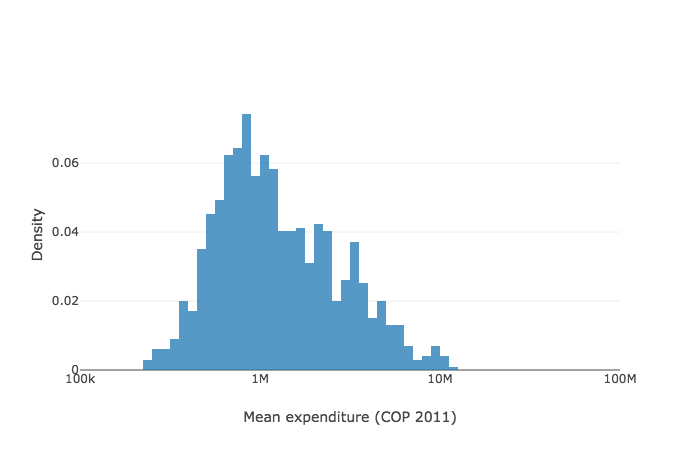}
\end{figure}

\begin{figure}
\caption{Wordclouds of observed categories with extreme values in expenditure distribution}
    \centering
    \begin{subfigure}[b]{0.55\textwidth}
       \includegraphics[width=1\linewidth]{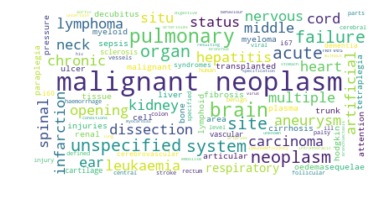}
       \caption{Above 95th percentile}
       \label{fig:Ng1} 
    \end{subfigure}
    
    \begin{subfigure}[b]{0.55\textwidth}
       \includegraphics[width=1\linewidth]{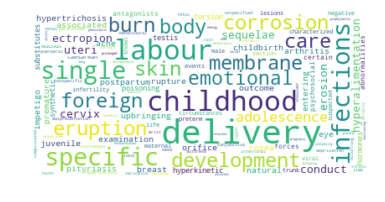}
       \caption{Below 5th percentile}
       \label{fig:Ng2}
    \end{subfigure}
\end{figure}

\subsection{Optimization}
Let $P_{k}$ a partition of the space of \textit{observed} three-character categories of size $k$. $p_{j} \in P_{k}, 1 \leq j \leq k$ is a  subset of three character categories of partition $P_{k}$. Let $d_{it}$ a dummy feature that indicates whether individual $i$ was diagnosed with $d$ at time $t$.  The dummy variable $D_{ijt}$ indicates affiliation of individual $i$ to diagnostic risk group $p_{j}$ at time $t$: $$D^{p_{j}}_{it} = \bigvee_{d \in p_{j}} \bigvee_{t\prime < t} d_{it\prime}$$  

In order to find \textit{optimal} diagnostic risk groups we split available observations into standard 80\% train and 20\% test splits. Table 1 presents descriptive statistics of train and test splits. We simulate multiple Markov chains with transitions distributions over partitions of the space of \textit{observed} three-character categories. We consider transition distributions that give uniform transition probability to all partitions at distance $j$, where $j \sim Poisson(\lambda)$ for different values of $k$. At every iteration of Metropolis-Hastings algorithm, we construct a linear hypothesis of expenditure using socio-demographic covariates and current three-character categories partitions and estimate out-of-sample mean absolute error. The linear hypothesis is constructed by OLS on the following specification:

\begin{equation*}
    y_{it} = \beta_{0} + \beta sex_{it} + \boldsymbol{\gamma \cdot age\_group_{it}} + \boldsymbol{\psi \cdot  residence\_group_{it}} + \boldsymbol{\phi \cdot D^{P_{k}}_{it}}
\end{equation*}

where $y_{it}$ denotes total value of health expenditure on individual $i$ at period $t$, $sex_{i}$ is a self explanatory dummy variable, $\boldsymbol{age\_group_{it}}$ is a column vector of dummy variables indicating affiliation of individual $i$ to age groups at time $t$, $\boldsymbol{residence\_group_{it}}$ is a column vector of dummy variables indicating affiliation of individual $i$ to residence groups at time $t$ and finally $\boldsymbol{D^{P_{k}}_{it}}$ is a column vector of dummy variables indicating affiliation of individual $i$ to diagnostic risk groups at time $t$:

$$ D^{P_{k}}_{it} = [D^{p_{1}}_{it},...,D^{p_{k}}_{it}]^{T}$$

Table 2 presents relevant statistics of the error distribution of every simulated Markov chain. Figure 7 shows the aggregate error distribution for varying partition size. Results suggest that, over the range of hyperparameters considered, increasing partition complexity improves mean prediction accuracy at the expense of increased variance in the error distribution. \textit{Optimal} diagnostic risk groups were found while iterating over a Markov chain of partitions of size 10 and considering neighbors at a distance distributed as a Poisson random variable with mean $\lambda=100$. We denote this \textit{optimal} risk groups as $MH_{10}$.

Results for low complexity partitions presented in figure 7 suggest that increasing partition size aids the learning process. However, as Figure 8 shows, an exploratory analysis of 500 iterations over high complexity partitions suggests that the learning potential of increasing $k$ is bounded. 

Working with high complexity partitions is expensive in terms of computational time. In order to reduce running time but still be able to benchmark our methodology, we consider finding \textit{optimal} risk groups over a Markov chain that has as initial seed risk groups constructed from expert criteria with $k=30$. We denote this risk groups as $E_{30}$. We also adjust parameter $T$ in order to have a higher rejection rate as shown in figure 9. We denote this high complexity \textit{optimal} risk groups as $MH_{30}$. A basic comparison of $E_{30}$ and $MH_{30}$ is presented in Appendix B. 

\newpage

\vspace*{\fill}
\vbox{
{\centering
    \captionof{table}{Feature means in training and test sets}
    \addvbuffer[10pt 10pt]{\begin{tabular}{lllr} \toprule
    & \multirow{2}{*}{Feature} & \multicolumn{2}{c}{Set}\\ \cmidrule(r){3-4}
    & & Train & Test \\ \midrule
    \multirow{1}{*}{Expenditure} & Expenditure in 2011 & 606,771 & 606,440 \\ \cmidrule(r){2-4}
    \multirow{11}{*}{Demographics} & Male & 0.443 & 0.442 \\
    & Age 0-1 & 0.013 & 0.014 \\
    & Age 2-4 & 0.053 & 0.052  \\
    & Age 5-18 & 0.202 & 0.200  \\
    & Age 19-44 & 0.423 & 0.423  \\
    & Age 45-49 & 0.070 & 0.069  \\
    & Age 50-54 & 0.060 & 0.060  \\
    & Age 55-59 & 0.048 & 0.048  \\
    & Age 60-64 & 0.038 & 0.039  \\
    & Age 65-69 & 0.029 & 0.030  \\
    & Age 70-74 & 0.023 & 0.024  \\
    & Age 74+ & 0.036 & 0.037  \\ 
    & Urban & 0.535 & 0.538 \\
    & Normal & 0.435 & 0.433 \\
    & Special & 0.029 & 0.028 \\
    
    \cmidrule(r){2-4}
    \multirow{10}{*}{\shortstack[l]{Random\\ ICD-10 categories}} & B23 - Human inmunodeficiency virus & $3.1*10^{-3}$ & $2.6*10^{-3}$ \\
    & C89 - Follicular lymphoma & $1.3*10^{-3}$ & $1.3*10^{-3}$ \\
    & E06 - Thyroiditis lymphoma & $7.6*10^{-3}$ & $6.3*10^{-3}$ \\
    & I51 - Ill-defined descriptions of heart disease & $9.9*10^{-3}$ & $9.1*10^{-3}$ \\
    & J38 - Diseases of vocal cords and larynx & $4.6*10^{-3}$ & $4.6*10^{-3}$ \\
    & K07 - Dentofacial anomalies & $8.4*10^{-2}$ & $9.0*10^{-2}$ \\
    & M68 - Disorders of synovium and tendon& $7.3*10^{-3}$ & $6.5*10^{-3}$ \\
    & Q24 - Congenital malformations of heart & $2.0*10^{-3}$ & $2.8*10^{-3}$ \\
    & S62 - Fracture at wrist and hand level  & $2.0*10^{-2}$ & $2.2*10^{-2}$ \\
    & Z13 - Special screening examination & $1.1*10^{-1}$ & $1.1*10^{-1}$ \\
    \bottomrule
    \end{tabular}}
    \par
}} 
\vspace*{\fill}
\newpage
\vspace*{\fill}
\vbox{
{\centering
    \captionof{table}{Error distribution over hyperparameter grid (thousands of COP)}
    \begin{tabular}{llccccc} \toprule
    \multirow{2}{*}{Partition size} & \multirow{2}{*}{Poisson mean} & \multicolumn{5}{c}{Result}\\ \cmidrule(r){3-7}
    & & Iterations & Minimum & Maximum & Mean & Std. deviation \\ \midrule
    \multirow{5}{*}{$k=2$} & $\lambda=5$ & 1000 & 730.1 & 733.8  & 732.0 & 0.61 \\ 
    & $\lambda=10$ & 1000 & 730.3 & 734.4 & 732.4 & 0.68\\
    & $\lambda=25$ & 1000 & 730.2 & 733.9 & 732.0 & 0.64\\
    & $\lambda=50$ & 1000& 730.1 & 734.3 & 732.2 & 0.67\\
    & $\lambda=100$ & 1000 & 729.4 & 734.1 & 732.1 & 0.69 \\
    \cmidrule(r){3-7}
    \multirow{5}{*}{$k=4$} & $\lambda=5$ & 1000 & 728.7 & 733.9 & 731.5 & 0.84 \\ 
    & $\lambda=10$ & 1000 & 729.3 & 734.1 & 731.8 & 0.89\\
    & $\lambda=25$ & 1000 & 729.2 & 734.7 & 731.8 & 0.84\\
    & $\lambda=50$ & 1000 & 729.0 & 735.0 & 731.7 & 1.00\\
    & $\lambda=100$ & 1000 & 728.6 & 735.2 & 732.0 & 1.00\\
    \cmidrule(r){3-7}
    \multirow{5}{*}{$k=6$} & $\lambda=5$ & 1000 & 728.2 & 736.5 & 731.6 & 1.35 \\ 
    & $\lambda=10$ & 1000 & 727.5 & 734.1 & 730.9 & 1.24\\
    & $\lambda=25$ & 1000 & 728.2 & 735.6 & 731.1 & 1.13\\
    & $\lambda=50$ & 1000 & 726.8 &  734.1 & 730.9 & 1.08 \\
    & $\lambda=100$ & 1000 & 727.2 & 735.2 & 731.2 & 1.18\\
    \cmidrule(r){3-7}
    \multirow{5}{*}{$k=8$} & $\lambda=5$ & 1000 & 726.5 & 735.0 & 730.5 & 1.70 \\ 
    & $\lambda=10$ & 1000 & 726.2 & 735.4 & 729.9 & 1.26\\
    & $\lambda=25$ & 1000 & 725.9 & 735.0 & 730.0 & 1.39 \\
    & $\lambda=50$ & 1000 & 725.8 & 734.7 & 730.2 & 1.36 \\
    & $\lambda=100$ & 1000 & 726.1 & 734.5 & 730.3 & 1.46 \\
    \cmidrule(r){3-7}
    \multirow{5}{*}{$k=10$} &$\lambda=5$ & 1000 & 724.3 & 733.6 & 729.1 & 1.47 \\ 
    & $\lambda=10$ & 1000 & 723.7 & 732.7 & 728.3 &  1.46 \\
    & $\lambda=25$ & 1000 & 725.1 & 734.6 &  729.5 & 1.73 \\
    & $\lambda=50$ & 1000 & 725.2 & 737.1 & 729.8  & 1.80 \\
    & $\lambda=100$ & 1000 & 723.5 & 734.9 & 729.6 & 1.79 \\
    \bottomrule
    \end{tabular}
    \par
}}
\vspace*{\fill}
\newpage
\begin{figure}
    \centering
    \includegraphics[scale=0.6]{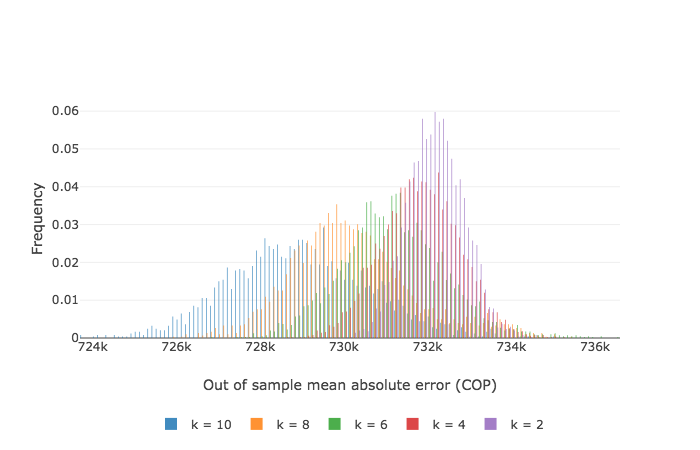}
    \caption{Error distribution over partition size}
\end{figure}

\begin{figure}
    \centering
    \includegraphics[scale=0.6]{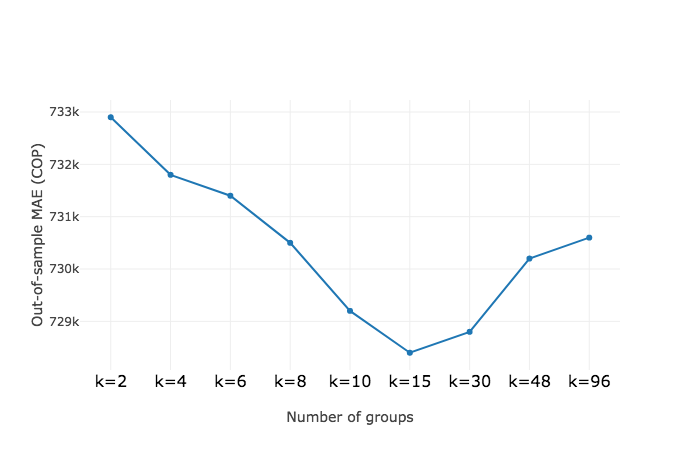}
    \caption{MAE over 500 iterations}
\end{figure}

\begin{figure}
    \centering
    \includegraphics[scale=0.6]{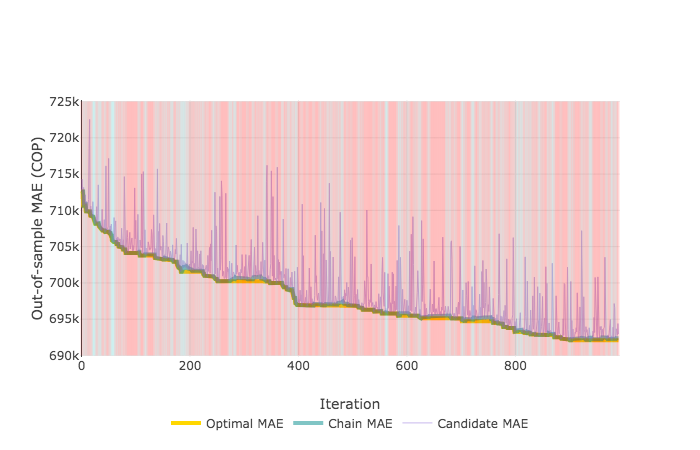}
    \caption{Markov chain for $MH_{30}$}
\end{figure}

\begin{figure}
    \centering
    \includegraphics[scale=0.6]{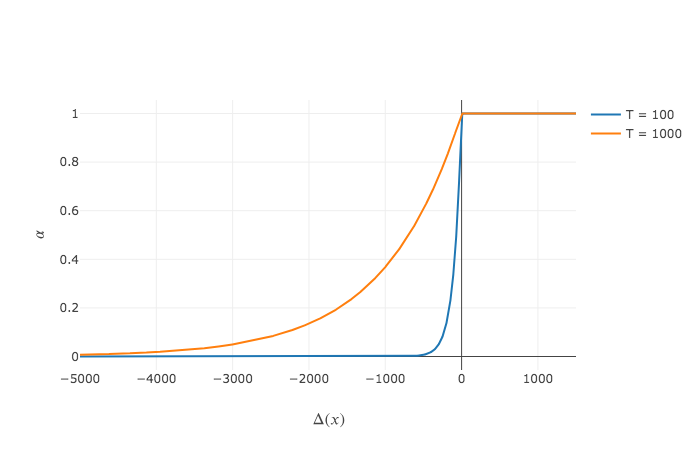}
    \caption{Comparison of energy function across temperatures}
\end{figure}

\subsection{Results}

In order to assess the relevance of the \textit{optimal} diagnostic risk groups found using our methodology we perform a 5-fold cross validation of three simple linear machines with different feature specifications:

\begin{equation}
    y_{it} = \beta_{0} + \beta sex_{it} + \boldsymbol{\psi \cdot  residence\_group_{it}}
\end{equation}

\begin{equation}
    y_{it} = \beta_{0} + \beta sex_{it} + \boldsymbol{\gamma \cdot age\_group_{it}} + \boldsymbol{\psi \cdot  residence\_group_{it}}
\end{equation}

\begin{equation}
    y_{it} = \beta_{0} + \beta sex_{it} + \boldsymbol{\gamma \cdot age\_group_{it}} + \boldsymbol{\psi \cdot  residence\_group_{it}} + \boldsymbol{\phi \cdot D^{E_{2}}_{it}}
\end{equation}

\begin{equation}
    y_{it} = \beta_{0} + \beta sex_{it} + \boldsymbol{\gamma \cdot age\_group_{it}} + \boldsymbol{\psi \cdot  residence\_group_{it}} + \boldsymbol{\phi \cdot D^{MH_{10}}_{it}}
\end{equation}

\begin{equation}
    y_{it} = \beta_{0} + \beta sex_{it} + \boldsymbol{\gamma \cdot age\_group_{it}} + \boldsymbol{\psi \cdot  residence\_group_{it}} + \boldsymbol{\phi \cdot D^{E_{30}}_{it}}
\end{equation}

\begin{equation}
    y_{it} = \beta_{0} + \beta sex_{it} + \boldsymbol{\gamma \cdot age\_group_{it}} + \boldsymbol{\psi \cdot  residence\_group_{it}} + \boldsymbol{\phi \cdot D^{MH_{30}}_{it}}
\end{equation}

Specification (1) considers expenditure as a linear function of residence and sex groups. Specification (2) considers expenditure as a linear function of sociodemographic covariates including age groups. Specification (3) considers expenditure as a linear function of sociodemographic covariates and affiliation to diagnostic risk groups with $k=2$ constructed from expert criteria as reported in Riascos (2018). Specification (4) considers expenditure as a linear function of sociodemographic covariates and affiliation to \textit{optimal} diagnostic risk groups $MH_{10}$. Specification (5) considers expenditure as a linear function of sociodemographic covariates and affiliation to diagnostic risk groups with $k=30$ constructed from expert criteria as reported in Riascos (2018). Finally, specification (6) considers expenditure as a linear function of sociodemographic covariates and affiliation to \textit{optimal} diagnostic risk groups $MH_{30}$.

Figure 11 and table 3 show the 5 fold cross-validation results for each specification considered over the full sample, the upper decile and the lower decile of the expenditure distribution. Inclusion of \textit{optimal} diagnostic risk groups improves expenditure prediction over the alternative specifications, in particular, complex expert risk groups. \\

\vbox{
{\centering
    \captionof{table}{Results of 5-fold cross validation over specifications (thousands of COP)}
    \begin{tabular}{ccccccccccc} \toprule
    \multirow{2}{*}{Spec.} & \multicolumn{3}{c}{Demographics} & \multicolumn{4}{c}{Partition} & \multicolumn{3}{c}{MAE}\\ \cmidrule(r){2-4} \cmidrule(r){5-8} \cmidrule(r){9-11}
    & $sex$ & $res$ & $age$ & $E_{2}$ & $MH_{10}$ & $E_{30}$ & $MH_{30}$ & $lower$ & $upper$ & $full$ \\ \midrule
    (1) & Y & Y & N & N & N & N & N & 600.7 & 3,755.1  & 777.0 \\
    (2) & Y & Y & Y & N & N & N & N & 494.5 & 3,514.1 & 738.1 \\
    (3) & Y & Y & Y & Y & N & N & N & 437.3 & 3,456.8 & 732.8 \\
    (4) & Y & Y & Y & N & Y & N & N & 366.9 & \textbf{3,390.7} & 729.1 \\
    (5) & Y & Y & Y & N & N & Y & N & 415.7 & 3,462.5 & 717.4 \\
    (6) & Y & Y & Y & N & N & N & Y & \textbf{350.1} & 3,400.7 & \textbf{697.2} \\
    \bottomrule
    \end{tabular}
    \par
}}

\vspace*{1 cm}

Table 4 shows the relative MAE among specifications. \textit{Optimal} risk groups constructed with Metropolis-Hastings reduce error in approximately $5\%$ with respect to current risk adjustment in the Colombian Healthcare System. Furthermore, these \textit{optimal} risk groups reduce error in approximately $2.6\%$ with respect to risk adjustment using expert risk groups with equal complexity. Improving risk adjustment by $5\%$ would have implied a redistribution of resources among insurers of approximately $\$400$ USD million in 2011.
This is significant amount of resources: between 2008 and 2013, system-wide resources that were redistributed among insurers by ex-post risk adjustment mechanisms amounted to $\$40$ USD million. \citep{cac}.

\vspace*{1 cm}

\vbox{
{\centering
    \captionof{table}{Relative MAE rate over full sample (\%)}
    \begin{tabular}{ccccccc} \toprule
    Spec & (1) & (2) & (3) & (4) & (5) & (6) \\ \midrule
    (1) & $0.00$ &  &  &  &   & \\
    (2) & $-5.00$ & $0.00$ &  &  &  & \\
    (3) & $-5.68$ & $-0.71$ & $0.00$ &  &  & \\
    (4) & $-6.16$ & $-1.21$ & $-0.50$ & $0.00$ &  & \\
    (5) & $-7.67$ & $-2.08$ & $-2.10$ & $-1.60$ &  $0.00$ & \\
    (6) & $-10.12$ & $-5.39$ & $-4.70$ & $-4.22$ &  $-2.66$ & $0.00$\\
    \bottomrule
    \end{tabular}
    \par
}}

\vspace*{1 cm}

Finally, table 5 presents aggregate prediction results over the test sample for selected specifications. Results suggest that linear models considered in this paper do not tend to underestimate or overestimate aggregate expenditure. 

\vspace*{1 cm}

\vbox{
{\centering
    \captionof{table}{Aggregate expenditure for select specifications}
    \begin{tabular}{cccccc} \toprule
    Spec & $n$ & $\sum y_{i}$ & $\sum \hat{y_{i}}$ & $ \frac{\sum\hat{y_{i}}}{n}$ & PR \\ \midrule
    (2) & $\num{1e5}$ & $\num{60.6e9}$ & $\num{61.1e9}$ & 611,990 & 1.01\\
    (5) & $\num{1e5}$ & $\num{60.6e9}$ & $\num{61.1e9}$ & 611,435 & 1.01\\
    (6) & $\num{1e5}$ & $\num{60.6e9}$ & $\num{60.8e9}$ & 608,533 & 1.00\\
    \bottomrule
    \end{tabular}
    \par
}}

\begin{figure}
    \caption{5-fold CV MAE for low complexity partitions}
    \begin{subfigure}{\linewidth}
      \centering 
      \includegraphics[scale=0.42]{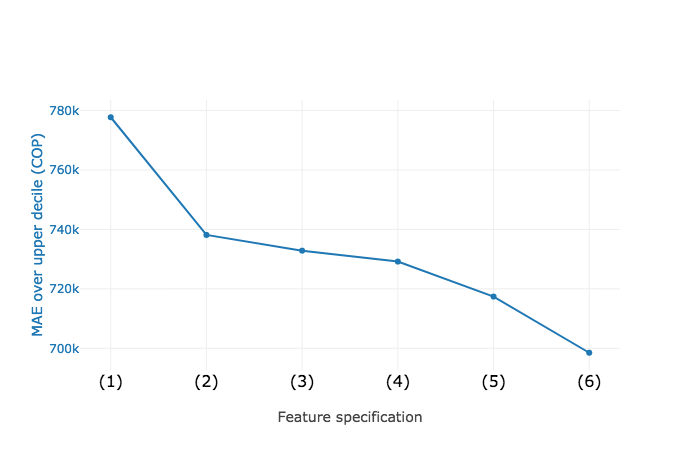}
      \caption{5-fold CV MAE over full sample}
    \end{subfigure}

    \begin{subfigure}[b]{\linewidth}
      \centering
      \includegraphics[scale=0.42]{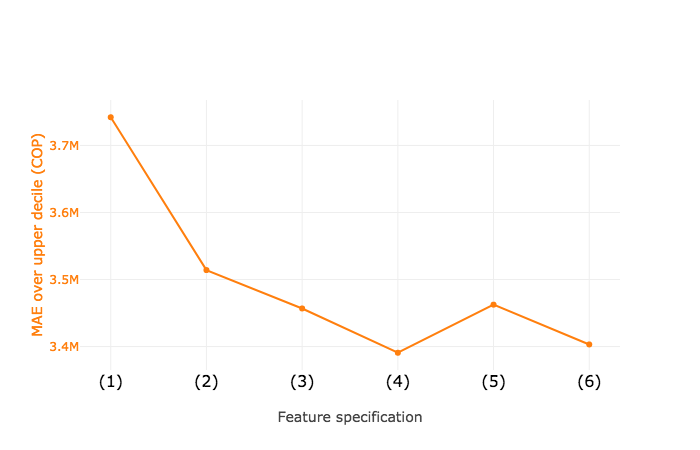}
      \caption{5-fold CV MAE over upper decile}
      \label{fig:blah1}
    \end{subfigure}
    
    \begin{subfigure}[b]{\linewidth}
      \centering
      \includegraphics[scale=0.42]{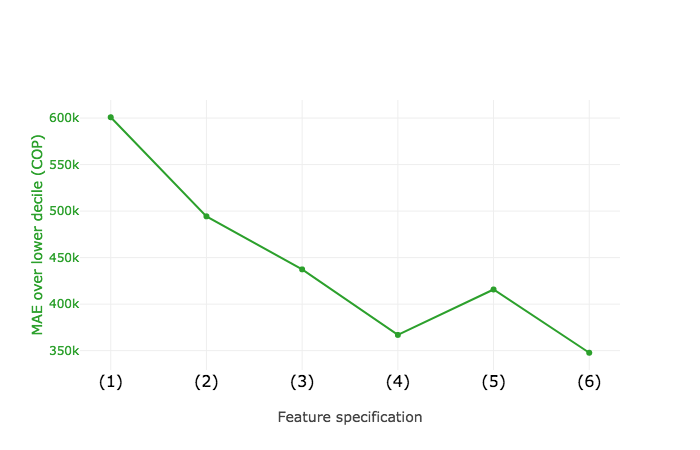}
      \caption{5-fold CV MAE over lower decile}
      \label{fig:blah2}
    \end{subfigure}
\end{figure}

\section{Discussion and Concluding Comments}

In this paper we propose a methodology to find optimal partitioning of categorical features for prediction based on the Metropolis-Hastings algorithm. Results for the problem of constructing diagnostic risk groups in the Colombian Health Sector from sample information show that our methodology outperforms common alternatives and has the potential to improve risk adjustment. Such improvement would reduce producer incentives to consumer \textit{selection} without encouraging \textit{production inefficiency} and thus it would effectively smooth \cite{Newhouse} production \textit{trade-off}. 

Results suggest that policy implications of adopting the proposed methodology for risk adjustment in Colombia could be significant. In principle, observed results imply that current efforts aimed at strengthening ex post risk adjustment are not efficient policy. However, further research is needed. First, a thorough evaluation should determine if results obtained in this paper using a sample cross section could be extended to the general population across time. Second, an empirical study of the complementarity of ex ante and ex post risk adjustment in the Colombian Healthcare System is needed. 

This paper opens multiple avenues for future research, both theoretical and applied. First, future research should consider finding a Metropolis-Hastings implementation that traverses the space of partitions of any (\textit{reasonable}) size. That is, at each iteration, consider the possibility of transitioning not only to neighboring partitions of the same size but also to neighboring partitions of lesser or greater size. This will probably require to redefine $Q(p,q)$ appropriately and while calculating $\alpha$ explicitly calculate $Q(x_{*},x_{i})$ and $Q(x_{i},x_{*})$.

Second, it would be desirable to have a clear expectation of convergence time for these type of Markov chains. This analysis would inform the critical decision of how many iterations are needed in practice to sample from the stationary distribution. 

Third, the results of this paper should be interpreted as exploratory, since only small samples of i) the Colombian Health Sector universe and ii) the hyperparameter space were considered. Working with panel data of a comprehensible proportion of enrollees and considering a expanded hyperparameter sample would add robustness to results. However, due to computational complexity, this scaling up would only be feasible on a distributed computing framework.

Finally, future research in diagnostic risk groups should consider further exploiting possible complementarities of expert knowledge and machine knowledge. As stated before, introducing expert knowledge to construct the initial distribution and constraining Metropolis-Hastings from reaching too far was instrumental in achieving high performance partitions. This is one of the many arrangements in which expert knowledge could inform MCMC.

\bigskip

\begin{appendices}
    \section{Algorithms}
    
    Metropolis-Hastings algorithm is remarkably simple:

    \begin{algorithm}[H]
                \textbf{Input}: $x_{0}$, $\pi(x)$ and $Q(\cdot)$\\
                \For{i=0,...,I}{
                    \text{Pick proposition $x_{*}$ using $Q(x_{i},\cdot)$}\\
                    \text{Calculate $\alpha = min\{1,\frac{\pi(x_{*})Q(x_{*},x_{i})}{\pi(x_{i})Q(x_{i},x_{*})}\}$}\\
                    With probability $\alpha$ set $x_{i+1} = x_{*}$, else $x_{i+1}=x_{i}$
                }
            \caption{MH pseudocode}
    \end{algorithm}
    
    In our problem, propositions $x_{0}, x_{1},...,x_{I}$ are partitions of $C$ of size $k$. We will denote by $x_{i}(c)$ the cluster of $c \in C$ under partition $x_{i}$.
    
    When using, in our algorithm, transitions to a partition at distance at most $j$ from $p$, we choose first the distance $j \sim Poisson(\lambda)$ at which $q$ will be. Then, the partition $q$ in $\{q \in P_{k} | d(p,q) = j\}$ is selected by choosing the $j$ elements in $C$ that must change cluster at random and then assigning at random the new cluster for
    each selected code. 
    
    The expression for $\pi(p)$ in 3.2.4, requires, in principle the computation of the expected loss at the current partition, $E[L(y,h_{p}(x,c))]$. Since the
    data distribution is not assumed to be known, this expectation must be approximated by its empirical version, which in our case, is the fitted sum of squares for the local linear model that is obtained by fitting a linear model to the observed responses over each member of $p$. The local linear model for $p$ and for the chosen transition candidate $q$ must be calculated at each step of the algorithm. 
    
    Our implementation of the algorithm with uniform transition probability among partitions at a random distance $j \sim Poisson(\lambda)$ follows closely:
    
    \begin{algorithm}[H]
                \textbf{Input}: $x_{0}$, $\pi(x)$, $Q(\cdot)$ and $k$\\
                \For{i=0,...,I-1}{
                    Let $x_{*} = x_{i}$\\
                    \text{Pick $j$ using $Poisson(\lambda)$}\\
                    Let $mem_{0}$ = \{\} \\
                    \For{i=0,...,j-1}{
                        \text{Pick $\widetilde{c_{i}}$ using uniform distribution over $C \setminus mem_{i}$}\\
                        \text{Let $mem_{i+1}$ = $mem_{i} \cup \{\widetilde{c_{i}}\}$} \\
                        \text{Pick $\widetilde{k_{i}}$ using uniform distribution over $\{1,...,k\} \setminus x_{i}(\widetilde{c_{i}})$}\\
                        Let $x_{*}(\widetilde{c_{i}}) = \widetilde{k_{}i}$\\
                    }
                    \text{Calculate $\alpha = min\{1,\frac{\pi(x_{*})}{\pi(x_{i})}\}$}\\
                    With probability $\alpha$ set $x_{i+1} = x_{*}$, else $x_{i+1}=x_{i}$
                    
                }
            \caption{MH for a fixed $k$ and random $j$ pseudocode}
    \end{algorithm}
    
To check algorithm 2 correctness, consider partitions $p$ and $q$ such that $D(p,q) = j$. Since $D(p,q) = j$ there exists a set of elements $E = \{c_{0},...,c_{j-1}\}$ such that $p(c_{i}) \neq q(c_{i})$, $0 \leq i \leq j$. The transition probability from $p$ to $q$ under algorithm 2 is:

$$Q(p,q) = P[Poisson(\lambda)=j]*\prod_{i=0}^{j-1} P[\widetilde{c_{i}} \in E \setminus mem_{i}]P[\widetilde{k_{i}}=q(c_{i})]$$

$$Q(p,q) = P[Poisson(\lambda)=j]*\prod_{i=0}^{j-1} \left(\frac{j-i}{n-i}\right)\left(\frac{1}{k-1}\right)$$

$$Q(p,q) = P[Poisson(\lambda)=j]*\left(\frac{1}{(k-1)^{j}}\right)*\prod_{i=0}^{j-1} \left(\frac{j-i}{n-i}\right)$$

$$Q(p,q) = P[Poisson(\lambda)=j]*\left(\frac{1}{(k-1)^{j}}\right)*\left(\frac{j}{n}*\frac{j-1}{n-1}*...*\frac{1}{n-(j-1)}\right)$$

$$Q(p,q) = P[Poisson(\lambda)=j]*\left(\frac{1}{(k-1)^{j}}\right)*\left(\frac{j!(n-j)!}{n!}\right)$$

$$Q(p,q) = P[Poisson(\lambda)=j]*\left(\frac{1}{(k-1)^{j}}\right)*\left(\frac{1}{{n\choose j}}\right)$$

$$ Q(p,q) =\frac{1}{{n\choose j}(k-1)^{j}}P[Poisson(\lambda)=j]$$

$$ Q(p,q) =\frac{1}{N_{p,j}}P[Poisson(\lambda)=j]$$

It is important to note that for a fixed $k$, $Q(p,q) = Q(q,p) \; \forall \; p,q \in P_{k}$ and therefore while calculating $\alpha$ there is no need to explicitly calculate $Q(x_{*},x_{i})$ and $Q(x_{i},x_{*})$.

\[
            Q(p,q) =
                \frac{1}{{n\choose j}(k-1)^{j}}P[Poisson(\lambda)=j]\\
        \]

\[
            Q(q,p) =
                \frac{1}{{n\choose j}(k-1)^{j}}P[Poisson(\lambda)=j]\\
        \]

\[
            Q(p,q) = Q(q,p)\\
        \]

Since $Q(p,q) = Q(q,p) \; \forall \; p,q \in P_{k}$ then 

$$\alpha = min\biggl\{1,\frac{\pi(x_{*})}{\pi(x_{i})}\biggr\}$$

$$\alpha = min\biggl\{1,\frac{exp\left(^\frac{-E[L(y,x_{*}(\cdot)]}{T}\right)}{exp\left(^\frac{-E[L(y,x_{i}(\cdot)]}{T}\right)}\biggr\}$$

$$\alpha = min\biggl\{1,exp\left(^\frac{E[L(y,x_{i}(\cdot)]-E[L(y,x_{*}(\cdot)]}{T}\right)\biggl\}$$

$$\alpha = min\biggl\{1,exp\left(^\frac{\mathcal{L}(x_{i})-\mathcal{L}(x_{*})}{T}\right)\biggr\}$$

$$\alpha = min\biggl\{1,exp\left(^\frac{\Delta(x)}{T}\right)\biggr\}$$

Parameter $T$ is commonly known as \textit{temperature} since for any given value of $\Delta(x)$ it determines the likelihood of \textit{jumping} to the candidate partitions. Bigger values of $T$ make jumps more likely and \textit{warm} the chain, while smaller values of $T$ make jumps less likely and \textit{chill} the chain.

Figure 10 shows $\alpha$ as a function of $\Delta(x)$ for different values of $T$. A value of $T = 1,000$ was used to encourage global exploration in chains with random seeds while $T = 100$ was used to encourage local exploration in chains with expert criteria informing seeds.

\section{Risk groups}

\vbox{
{\centering
    \captionof{table}{Comparison of expert and \textit{optimal} risk groups}
    \begin{tabular}{lccc} \toprule
    Group & Expert tag & Size under $E^{30}$ & Size under $MH^{30}$ \\ \midrule
    0 & SIDA-VIH & 8 & 27\\
    1 & TUBERCULOSIS & 6 & 37\\
    2 & CANCER - OTRO CANCER & 24 & 29\\
    3 & CANCER - ORGANOS DIGESTIVOS & 3 & 34\\
    4 & CANCER - ORGANOS RESPIRATORIOS & 2 & 23\\
    5 & CANCER - MELANOMA Y PIEL & 2 & 27\\
    6 & CANCER - MAMA & 2 & 27\\
    7 & CANCER - OTROS GENITALES FEMENINOS & 6 & 37\\
    8 & CANCER - CERVIX INVASIVO & 1 & 38\\
    9 & CANCER - GENITALES MASCULINOS & 2 & 19\\
    10 & CANCER - TEJIDOS LINFATICOS & 9 & 43\\
    11 & CANCER - CERVIX IN SITU& 3 & 30\\
    12 & AUTOINMUNE & 6 & 30\\
    13 & DIABETES & 13 & 31\\
    14 & SINDROMES CONVULSIVOS & 3 & 38\\
    15 & CARDIOVASCULAR - OTRA & 40 & 36\\
    16 & CARDIOVASCULAR - HIPERTENSION & 4 & 31\\
    17 & PULMONAR LARGA DURACION & 10 & 22\\
    18 & ASMA & 2 & 32\\
    19 & ARTRITIS PIOGENAS Y REACTIVAS & 3 & 22\\
    20 & ARTRITIS & 8 & 32\\
    21 & ARTROSIS & 5 & 24\\
    22 & RENAL & 7 & 29\\
    23 & RENAL - LARGA DURACION & 1 & 32\\
    24 & TRANSPLANTE& 2 & 17\\
    25 & INSUFICIENCIA RENAL & 1 & 30\\
    26 & INSUFICIENCIA RENAL CRONICA& 1 & 9\\
    27 & ANOMALIAS GENETICAS& 32 & 26\\
    28 & CANCER - TERAPIA CANCER & 3 & 32\\
    29 & OTRAS & 786 & 151\\
    \bottomrule
    \end{tabular}
    \par
}}

\end{appendices}

\bigskip
\bigskip
  
\bibliography{bibliography.bib}

\end{document}